\documentclass[11pt]{article}

\usepackage{coling2020}
\usepackage{times}
\usepackage{url}
\usepackage{latexsym}
\usepackage[linesnumbered,ruled,vlined]{algorithm2e}
\usepackage{xcolor}
\usepackage{graphicx}
\usepackage[]{lineno} 
\usepackage{pbox}
\usepackage{makecell}
\usepackage{multirow}

\definecolor{ultramarine}{rgb}{0.07, 0.04, 0.56}
\usepackage{hyperref}
\hypersetup{
    colorlinks=true,
    linkcolor=ultramarine,
    citecolor=ultramarine,
    urlcolor=ultramarine,
}

\definecolor{mypink1}{rgb}{0.858, 0.188, 0.478}
\definecolor{mygreen}{rgb}{0.258, 0.888, 0.178}

\title{Constructing A Multi-hop QA Dataset for Comprehensive Evaluation of Reasoning Steps}

\colingfinalcopy 

\newcommand{\sy}[0]{\clubsuit}
\newcommand{\sz}[0]{\diamondsuit}
\newcommand{\sw}[0]{\heartsuit}

\author{Xanh Ho$^{\sw\sy}$, Anh-Khoa Duong Nguyen$^\sz$, Saku Sugawara$^\sy$, Akiko Aizawa$^{\sw\sy}$ \\
$^\sw$ The Graduate University for Advanced Studies, Kanagawa, Japan \\ 
$^\sy$ National Institute of Informatics, Tokyo, Japan \\
$^\sz$ National Institute of Advanced Industrial Science and Technology, Tokyo, Japan \\
{\tt \{xanh, saku, aizawa\}@nii.ac.jp} \\
{\tt khoa.duong@aist.go.jp} 
}

\date{}

\usepackage{tikz}
\def\checkmark{\tikz\fill[scale=0.4](0,.35) -- (.25,0) -- (1,.7) -- (.25,.15) -- cycle;}

\newcommand{\myedit}[1]{\textcolor{black}{#1}}

\begin{document}
\maketitle

\begin{abstract}
A multi-hop question answering (QA) dataset aims to test reasoning and inference skills by requiring a model to read multiple paragraphs to answer a given question.
However, current datasets do not provide a complete explanation for the reasoning process from the question to the answer.
Further, previous studies revealed that many examples in existing multi-hop datasets do not require multi-hop reasoning to answer a question.
In this study, we present a new multi-hop QA dataset, called 2WikiMultiHopQA, which uses structured and unstructured data.
In our dataset, we introduce the evidence information containing a reasoning path for multi-hop questions.
The evidence information has two benefits: (i) providing a comprehensive explanation for predictions and (ii) evaluating the reasoning skills of a model.
We carefully design a pipeline and a set of templates when generating a question--answer pair that guarantees the multi-hop steps and the quality of the questions.
We also exploit the structured format in Wikidata and use logical rules to create questions that are natural but still require multi-hop reasoning.
Through experiments, we demonstrate that our dataset is challenging for multi-hop models and it ensures that multi-hop reasoning is required.
\end{abstract}

\section{Introduction}
\label{intro}

\blfootnote{
    %
    \hspace{-0.65cm}  
    This work is licensed under a Creative Commons 
    Attribution 4.0 International License.
    License details:
    \url{http://creativecommons.org/licenses/by/4.0/}.
}

Machine reading comprehension (MRC) aims at teaching machines to read and understand given text.
Many current models~\cite{cite_model_bert,model_robert,model_xlnet} have defeated humans on the performance of SQuAD~\cite{data_squad1,data_squad2}, as shown on its leaderboard\footnote{\url{https://rajpurkar.github.io/SQuAD-explorer/}}. 
However, such performances do not indicate that these models can completely understand the text.
Specifically, using an adversarial method,~\newcite{eval_adversarial} demonstrated that the current models do not precisely understand natural language.
Moreover,~\newcite{eval_fewwords} demonstrated that many datasets contain a considerable number of easy instances that can be answered based on the first few words of the questions.

Multi-hop MRC datasets require a model to read and perform multi-hop reasoning over multiple paragraphs to answer a question.
Currently, there are four multi-hop datasets over textual data: ComplexWebQuestions~\cite{data_ComplexWebQuestions}, QAngaroo~\cite{data_Qangaroo}, HotpotQA~\cite{data_Hotpotqa}, and $\mathrm{R^4C}$~\cite{data_r4c}.
The first two datasets were created by incorporating the documents (from Web or Wikipedia) with a knowledge base (KB).
Owing to their building procedures, these datasets have no information to explain the predicted answers. 
Meanwhile, the other two datasets were created mainly based on crowdsourcing.
In HotpotQA, the authors introduced the sentence-level supporting facts (SFs) information that are used to explain the predicted answers. 
However, as discussed in~\newcite{data_r4c}, the task of classifying sentence-level SFs is a binary classification task that is incapable of evaluating the reasoning and inference skills of the model.
Further, data analyses~\cite{data_analysis_chen,data_analysis_sewon} revealed that many examples in HotpotQA do not require multi-hop reasoning to solve.

Recently, to evaluate the internal reasoning of the reading comprehension system,~\newcite{data_r4c} proposed a new dataset $\mathrm{R^4C}$ that requires systems to provide an answer and derivations.
A derivation is a semi-structured natural language form that is used to explain the answers.
$\mathrm{R^4C}$ is created based on HotpotQA and has 4,588 questions.
However, the small size of the dataset implies that the dataset cannot be used as a multi-hop dataset with a comprehensive explanation for training end-to-end systems.

\begin{figure}[htp]
    \includegraphics[scale=0.51]{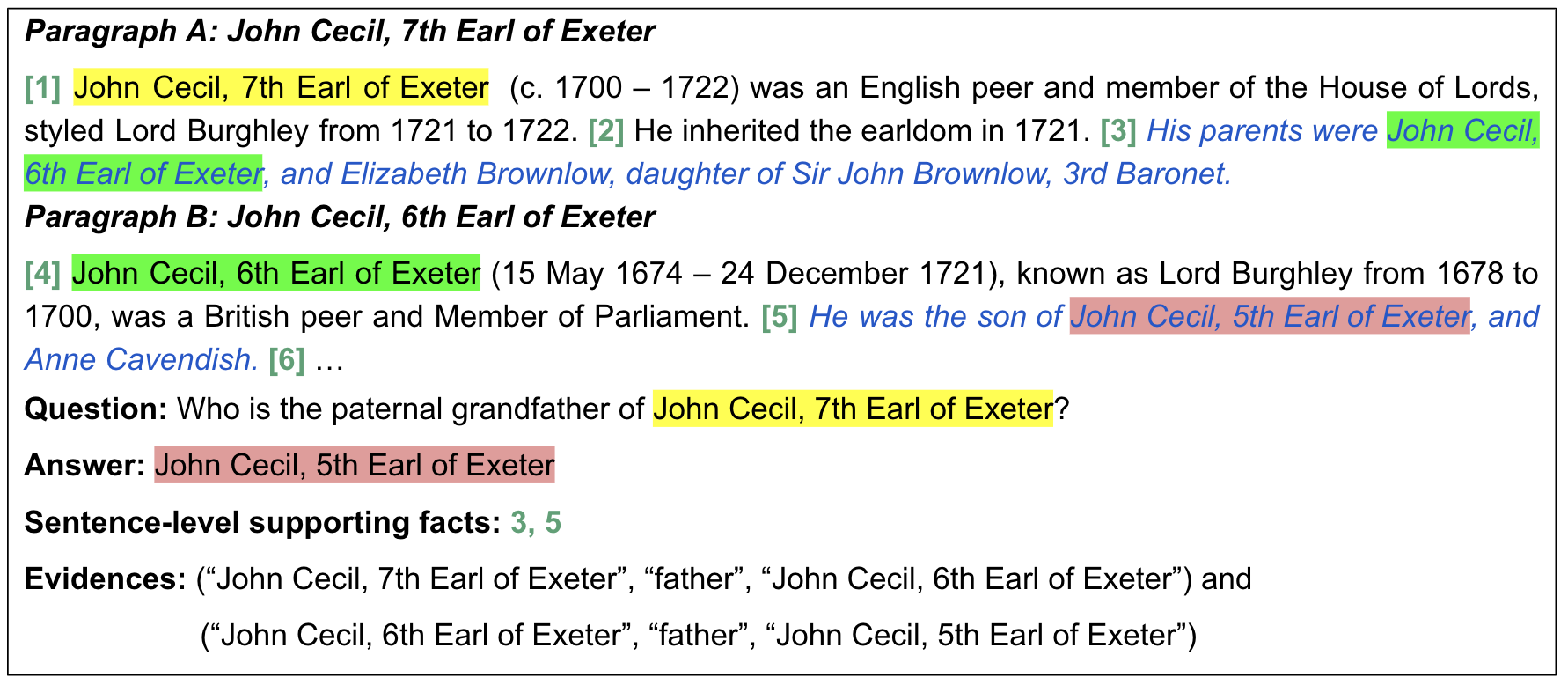}
    \caption{Example of an inference question in our dataset. The difference between our dataset and HotpotQA is the evidence information that explains the reasoning path.}
    \label{fig:ex_inference_ques}
\end{figure}

In this study, we create a large and high quality multi-hop dataset 2WikiMultiHopQA\footnote{2Wiki is a combination of Wikipedia and Wikidata.} with a comprehensive explanation by combining structured and unstructured data.
To enhance the explanation and evaluation process when answering a multi-hop question on Wikipedia articles, we introduce new information in each sample, namely \textit{evidence} that contains comprehensive and concise information to explain the predictions.
Evidence information in our dataset is a set of triples, where each triple is a structured data \textit{(subject entity, property, object entity)} obtained from the Wikidata (see Figure~\ref{fig:ex_inference_ques} for an example).

Our dataset has four types of questions: comparison, inference, compositional, and bridge comparison.
All questions in our dataset are created by using a set of predefined templates.
\newcite{data_analysis_sewon} classified the comparison questions in HotpotQA in three types: multi-hop, context-dependent multi-hop, and single-hop.
Based on this classification, we removed all templates in our list that make questions become single-hop or context-dependent multi-hop to ensure that our comparison questions and bridge-comparison questions are multi-hop.
We carefully designed a pipeline to utilize the intersection information between the summary\footnote{Another name is ``short description''; The short description at the top of an article that summarizes the content. See also \url{https://en.wikipedia.org/wiki/Wikipedia:Short_description}} of Wikipedia articles and Wikidata and have a special treatment for each type of question that guarantees multi-hop steps and the quality of the questions.
Further, by utilizing the logical rule information in the knowledge graph, such as $father(a, b) \wedge father(b, c) \Rightarrow grandfather(a, c)$, we can create more natural questions that still require multi-hop reasoning.

We conducted two different evaluations on our dataset: difficulty and multi-hop reasoning of the dataset.
To evaluate the difficulty, we used a multi-hop model to compare the performance of HotpotQA and our dataset.
\myedit{Overall, the results from our dataset are lower than those observed in HotpotQA, while human scores are comparable on both datasets.}
This suggests that the number of difficult questions in our dataset is greater than that in HotpotQA.
Similar to~\newcite{data_analysis_sewon}, we used a single-hop BERT model to test the multi-hop reasoning in our dataset.
The result of our dataset is lower than the result of HotpotQA by 8.7 F1, indicating that a lot of examples in our dataset require multi-hop reasoning to be solved.
Through experiments, we confirmed that although our dataset is generated by hand-crafted templates and the set of predefined logical rules, it is challenging for multi-hop models and requires multi-hop reasoning.

In summary, our main contributions are as follows:
(1) We use Wikipedia and Wikidata to create a large and high quality multi-hop dataset that has comprehensive explanations from question to answer.
(2) We provide new information in each sample---evidence information useful for interpreting the predictions and testing the reasoning and inference skills of the model.
(3) We use logical rules to generate a simple natural question but still require the model to undertake multi-hop reasoning when answering a question.
The full dataset, baseline model, and all information that we used when constructing the dataset are available at \url{https://github.com/Alab-NII/2wikimultihop}.

\section{Task Overview}
\label{overview}

\subsection{Task Formalization and Metrics}
\label{task_evaluation}
We formulated (1) answer prediction, (2) sentence-level SFs prediction, and (3) evidence generation tasks as follows:
\begin{itemize}
    \item Input: a question $Q$ and a set of documents $D$.

    \item Output: (1) find an answer $A$ (a textual span in $D$) for $Q$, (2) find a set of sentence-level SFs (sentences) in $D$ that a model used to answer $Q$, and (3) generate a set of evidence $E$ which consists of triples that describes the reasoning path from $Q$ to $A$.
\end{itemize}

We evaluate the three tasks by using two evaluation metrics: exact match (EM) and F1 score.
Following previous work~\cite{data_Hotpotqa}, to assess the entire capacity of the model, we introduced joint metrics that combine the evaluation of answer spans, sentence-level SFs, and evidence as follows:



\begin{equation}
    Joint\: F1 = \frac{2P^{joint} R^{joint}}{P^{joint} + R^{joint}}
\end{equation}

where $P^{joint} = P^{ans}P^{sup}P^{evi}$ and 
$R^{joint} = R^{ans}R^{sup}R^{evi}$.
$(P^{ans}$, $R^{ans})$, $(P^{sup}, R^{sup})$, and $(P^{evi}, R^{evi})$ denote the precision and recall of the answer spans, sentence-level SFs, and evidence, respectively.
Joint EM is 1 only when all the three tasks obtain an exact match or otherwise 0.

\subsection{Question Types}

In our dataset, we have the following four types of questions:
(1) comparison, (2) inference, (3) compositional, and (4) bridge comparison. 
The inference and compositional questions are the two subtypes of the bridge question which comprises a bridge entity that connects the two paragraphs~\cite{data_Hotpotqa}.

\begin{enumerate}
    \item \textbf{Comparison question} is a type of question that compares two or more entities from the same group in some aspects of the entity~\cite{data_Hotpotqa}. 
    For instance, a comparison question compares two or more people with the \textit{date of birth} or \textit{date of death} (e.g., \textit{Who was born first, Albert Einstein or Abraham Lincoln?}).

    \item \textbf{Inference question} is created from the two triples $(e, r_1, e_1)$ and $(e_1, r_2, e_2)$ in the KB. 
    We utilized the logical rule to acquire the new triple $(e, r, e_2)$, where $r$ is the inference relation obtained from the two relations $r_1$ and $r_2$.
    A question--answer pair is created by using the new triple $(e, r, e_2)$,
    its question is created from $(e, r)$ and its answer is $e_2$.
    \myedit{For instance, using two triples \textit{(Abraham Lincoln, mother, Nancy Hanks Lincoln)} and \textit{(Nancy Hanks Lincoln, father, James Hanks)}, 
    we obtain a new triple \textit{(Abraham Lincoln, maternal grandfather, James Hanks)}. A question is: \textit{Who is the maternal grandfather of Abraham Lincoln?} An answer is \textit{James Hanks}  (Section~\ref{data_collection:inference}).}

    \item \textbf{Compositional question} is created from the two triples $(e, r_1, e_1)$ and $(e_1, r_2, e_2)$ in the KB. 
    Compared with inference question, the difference is that no inference relation $r$ exists from the two relations $r_1$ and $r_2$.
    \myedit{For instance, there are two triples \textit{(La La Land, distributor, Summit Entertainment)} and \textit{(Summit Entertainment, founded by, Bernd Eichinger)}. 
    There is no inference relation $r$ from the two relations \textit{distributor} and \textit{founded-by}. In this case, a question is created from the entity $e$ and the two relations $r_1$ and $r_2$: \textit{Who is the founder of the company that distributed La La Land film?} An answer is the entity $e_2$ of the second triple: \textit{Bernd Eichinger} (Section~\ref{data_collection:compositional}).}

    \item \textbf{Bridge-comparison question} is a type of question that combines the bridge question with the comparison question.
    It requires both finding the bridge entities and doing comparisons to obtain the answer. 
    \myedit{For instance, instead of directly compare two films, we compare the information of the directors of the two films, e.g., \textit{Which movie has the director born first, La La Land or Tenet?}
    To answer this type of question, the model needs to find the bridge entity that connects the two paragraphs, one about the film and one about the director, to get the date of birth information.
    Then, making a comparison to obtain the final answer.}
    
\end{enumerate}

\section{Data Collection}
\label{data_collection}

\subsection{Wikipedia and Wikidata}
In this study, we utilized both text descriptions from Wikipedia\footnote{\url{https://www.wikipedia.org}} and a set of statements from Wikidata to construct our dataset.
%
We used only a summary from each Wikipedia article as a paragraph that describes an entity.
Wikidata\footnote{\url{https://www.wikidata.org}} is a collaborative KB that stores data in a structured format.
Wikidata contains a set of statements (each statement includes property and an object entity) to describe the entity.
There is a connection between Wikipedia and Wikidata for each entity.
From Wikidata, we can extract a triple $(s, r, o)$, where $s$ is a subject entity, $r$ is a property or relation, and $o$ is an object entity.
A statement for the entity $s$ is $(r, o)$.
An object entity can be another entity or the date value.
We categorized all entities based on the value of the property \textit{instance of} in Wikidata (Appendix~\ref{appendix_data_process}).

\subsection{Dataset Generation Process}
\label{data_collection:compositional}
\label{data_collection:inference}
Generating a multi-hop dataset in our framework involves three main steps: (1) create a set of templates, (2) generate data, and (3) post-process generated data.
After obtaining the generated data, we used a model to split the data into \textit{train}, \textit{dev}, and \textit{test} sets.

\paragraph{(1) Create a Set of Templates:}
For the comparison question, first, we used Spacy\footnote{\url{https://spacy.io/}} to extract named entity recognition (NER) tags and labels for all comparison questions in the train data of HotpotQA (17,456 questions). 
Then, we obtained a set of templates $L$ by replacing the words in the questions with the labels obtained from the NER tagger.
We manually created a set of templates based on $L$ for entities in the top-50 most popular entities in Wikipedia.
We focused on a set of specific properties of each entity type (Appendix~\ref{appendix_comparison}) in the KB.
We also discarded all templates that made questions become single-hop or context-dependent multi-hop as discussed in~\newcite{data_analysis_sewon}. 
Based on the templates of the comparison question, we manually enhanced it to create the templates for bridge-comparison questions (Appendix~\ref{appendix_combine}).
We manually created all templates for inference and compositional questions (Appendix~\ref{appendix_inference} and~\ref{appendix_composite}).

For the inference question, we utilized logical rules in the knowledge graph to create a simple question but still require multi-hop reasoning.
Extracting logical rules is a task in the knowledge graph wherein the target makes the graph complete.
We observe that logical rules, such as $spouse(a, b) \wedge mother(b, c) \Rightarrow mother\_in\_law(a, c)$, can be used to test the reasoning skill of the model.
Based on the results of the AMIE model~\cite{logical_rule_amie}, we manually checked and verified all logical rules to make it suitable for the Wikidata relations.
We obtained 28 logical rules (Appendix~\ref{appendix_inference}).

\paragraph{(2) Generate Data:}
From the set of templates and all entities' information, we generated comparison questions as described in Algorithm~\ref{al:generate_compare_data} (Appendix~\ref{generate_data}).
For each entity group, we randomly selected two entities: $e_1$ and $e_2$.
Subsequently, we obtained the set of statements of each entity from Wikidata. 
Then, we processed the two sets of statements to obtain a set of mutual relations ($M$) between two entities.
We then acquired the Wikipedia information for each entity.
For each relation in $M$, for example, a relation $r_1$, we checked whether we can use this relation.
Because our dataset is a span extraction dataset, the answer is extracted from the Wikipedia article of each entity.
With relation $r_1$, we obtained the two values $o_1$ and $o_2$ from the two triples $(e_1, r_1, o_1)$ and $(e_2, r_1, o_2)$ of the two entities, respectively.
The requirement here is that the value $o_1$ must appear in the Wikipedia article for the entity $e_1$, which is the same condition for the second entity $e_2$.

When all information passed the requirements, we generated a question--answer pair that includes a question $Q$, a context $C$, the sentence-level SFs $SF$, the evidence $E$, and an answer $A$.
$Q$ is obtained by replacing the two tokens \textit{\#name} in the template by the two entity labels.
$C$ is a concatenation of the two Wikipedia articles that describe the two entities.
$E$ is
the two triples $(e_1, r_1, o_1)$ and $(e_2, r_1, o_2)$.
$SF$ is a set of sentence indices where the values $o_1$ and $o_2$ are extracted.
Based on the type of questions, we undertake comparisons and obtain the final answer $A$.

\begin{figure}[htp]
\begin{center}
    \includegraphics[scale=0.4]{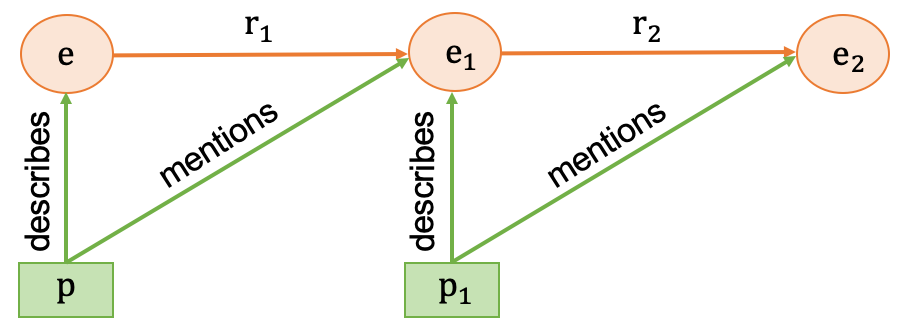}
\end{center}
    \caption{The Requirements for bridge questions in our dataset.}
    \label{fig:2_hop_requirement}
\end{figure}

We generated bridge questions as described in Algorithm~\ref{al:generate_brdige_data} (Appendix~\ref{generate_data}).
For each entity group, we randomly selected an entity $e$ and then obtained a set of statements of the entity from Wikidata.
Subsequently, based on the first relation information in $R$ (the set of predefined relations), we filtered the set of statements to obtain a set of 1-hop $H_1$.
Next, for each element in $H_1$, we performed the same process to obtain a set of 2-hop $H_2$, each element in $H_2$ is a tuple $(e, r_1, e_1, r_2, e_2)$.
For each tuple in $H_2$, we obtained the Wikipedia articles for two entities $e$ and $e_1$.
Then, we checked the requirements to ensure that this sample can become a multi-hop dataset.
For instance, the two paragraphs $p$ and $p_1$ describe for $e$ and $e_1$, respectively (see Figure~\ref{fig:2_hop_requirement}).
The bridge entity requirement is that $p$ must mention $e_1$.
The span extraction answer requirement is that $p_1$ must mention $e_2$.
The 2-hop requirements are that $p$ must not contain $e_2$ and $p_1$ must not contain $e$.
Finally, we obtained $Q$, $C$, $SF$, $E$, and $A$ similarly to the process in comparison questions.

\paragraph{(3) Post-process Generated Data:}
We randomly selected two entities to create a question when generating the data; therefore, a large number of \textit{no} questions exist in the \textit{yes/no} questions.
We performed post-processing to finalize the dataset that balances the number of \textit{yes} and \textit{no} questions.
Questions could have several true answers in the real world.
To ensure one sample has only one answer, we discarded all ambiguous cases in the dataset (Appendix~\ref{post_process_data}).

\paragraph{Collect Distractor Paragraphs:}
Following~\newcite{data_Hotpotqa} and~\newcite{data_analysis_sewon}, we used bigram tf-idf ~\cite{chen_bigram} to retrieve the top-50 paragraphs from Wikipedia that are most similar to the question.
Then, we used the entity type of the two gold paragraphs (four gold paragraphs for bridge-comparison question) to select the top-8 paragraphs (top-6 for bridge-comparison question) and considered it as a set of distractor paragraphs.
We shuffled the 10 paragraphs (including gold and distractor paragraphs) and obtained a context.

\paragraph{Dataset Statistics (A Benchmark Setting):}
We used a single-hop model (Section~\ref{equaluate_multihop}) to split the \textit{train}, \textit{dev}, and \textit{test} sets. 
We conducted five-fold cross-validation on all data.
The average F1 score of the model is 86.7\%.
All questions solved by the single-hop model are considered as a \textit{train-medium} subset.
The rest was split into three subsets: \textit{train-hard}, \textit{dev}, and \textit{test} (balancing the number of different types of questions in each subset).
Statistics of the data split can be found in Table~\ref{tab:data_split}.
We used \textit{train-medium} and \textit{train-hard} as the training data in our dataset.

\begin{table}[!htb]
    \begin{minipage}[t]{.4\linewidth}
      \centering
    \begin{tabular}{llr}  \Xhline{2\arrayrulewidth}
      \textbf{Name}  & \textbf{Split} & \textbf{\#Examples} \\\hline
      train-medium  & train & 154,878 \\
      train-hard & train & 12,576 \\
      dev & dev & 12,576\\
      test  & test & 12,576\\\hline
      Total &  & 192,606\\ \Xhline{2\arrayrulewidth}
    \end{tabular}
    \caption{Data statistics.}
    \label{tab:data_split}
    \end{minipage}%
    \begin{minipage}[t]{.05\linewidth}
    \end{minipage}
    \begin{minipage}[t]{.6\linewidth}
      \centering
    \begin{tabular}{lrrr}  \Xhline{2\arrayrulewidth}
      \textbf{Type of Q} & \textbf{\#Examples} & \textbf{\#Avg. Q} & \textbf{\#Avg. A} \\\hline
      Comparison & 57,989 & 11.97 & 1.58 \\
      Inference & 7,478 & 8.41 & 3.15 \\
      Compositional & 86,979 & 11.43 & 2.05 \\
      Bridge-comparison & 40,160 & 17.01 & 2.01 \\\hline
      Total & 192,606 & 12.64  & 1.94  \\ \Xhline{2\arrayrulewidth}
    \end{tabular}
    \caption{Question and answer lengths across the different type of questions. \textbf{Q} is the abbreviation for ``question'', and \textbf{A} is for ``answer''.}
    \label{tab:data_statistic_detail}
    \end{minipage} 
\end{table}

\section{Data Analysis}
\label{data_analysis}

\paragraph{Question and Answer Lengths}
\myedit{
We quantitatively analyze the properties of questions and answers for each type of question in our dataset. The statistics of the dataset are presented in Table~\ref{tab:data_statistic_detail}.
The compositional question has the greatest number of examples, and the inference question has the least number of examples.
To ensure one question has only one possible answer, we used the information from Wikidata and removed many inference questions that may have more than one answer.
The average question length of the inference questions is the smallest because they are created from one triple.
The average question length of the bridge-comparison questions is the largest because it combines both bridge question and comparison question.
The average answer lengths of comparison and bridge-comparison questions are smaller than inference and compositional questions.
This is because there are many \textit{yes/no} questions in the comparison questions.
}

\paragraph{Multi-hop Reasoning Types}

\begin{table}[htp]
\centering
\begin{tabular}{p{3cm}p{12.2cm}}
\Xhline{2\arrayrulewidth}
\textbf{Reasoning Type} & \textbf{Example} \\ \hline

\multirow{3}{3cm}{Comparison question: comparing two entities} 
 & \textbf{Paragraph A:} {\color{mygreen} Theodor Haecker} (June 4, 1879 - April 9, 1945) was a \ldots \\
& 
\textbf{Paragraph B:} {\color{orange} Harry Vaughan Watkins} (10 September 1875 – 16 May 1945) was a Welsh rugby union player \ldots \\
& 
\textbf{Q:} Who lived longer, {\color{mygreen} Theodor Haecker} or {\color{orange} Harry Vaughan Watkins}?
 \\ \hline
 
 \multirow{3}{3cm}{Compositional question: inferring the bridge entity to find the answer} 
  & \textbf{Paragraph A:} {\color{orange} Versus} (Versace) is the diffusion line of Italian \ldots, a gift by the founder {\color{blue} Gianni Versace}  to his sister, Donatella Versace. \ldots  \\
 & \textbf{Paragraph B:} {\color{blue} Gianni Versace}  \ldots Versace was {\color{mypink1} shot} and killed outside \ldots \\
 & \textbf{Q:} Why did the founder of {\color{orange} Versus} die?  \\ \hline

  \multirow{3}{3cm}{Inference question: using logical rules and inferring the bridge entity} 
 & \textbf{Paragraph A: } {\color{orange} Dambar Shah} (? – 1645) was the king of the Gorkha Kingdom \ldots
 He was the father of {\color{blue} Krishna Shah}.  \ldots \\
 & \textbf{Paragraph B:} {\color{blue} Krishna Shah} (? – 1661) \ldots
He was the father of {\color{mypink1} Rudra Shah}. \\
 & \textbf{Q:} Who is the grandchild of {\color{orange} Dambar Shah}? \\ \hline
 
  \multirow{5}{3cm}{Bridge-comparison question: inferring the bridge entity and doing comparisons} 
 & \textbf{Paragraph A:} {\color{orange} FAQ: Frequently Asked Questions} is a feature-length dystopian movie, written and directed by {\color{blue} Carlos Atanes} and released in 2004. \ldots \\
 & \textbf{Paragraph B:} {\color{mygreen} The Big Money} \ldots directed by {\color{blue} John Paddy Carstairs}  \ldots \\
  & \textbf{Paragraph C:} {\color{blue} Carlos Atanes} is a {\color{mypink1} Spanish} film director \ldots \\
 & \textbf{Paragraph D:} {\color{blue} John Paddy Carstairs} was a prolific  {\color{mypink1} British} film director \ldots \\
 & \textbf{Q:}  Are both director of film {\color{orange} FAQ: Frequently Asked Questions} and director of film {\color{mygreen} The Big Money} from the same country? \\ 
 
\Xhline{2\arrayrulewidth}
\end{tabular}
\caption{\label{tab-multi-hop_type} Types of multi-hop reasoning in our dataset. }
\vspace{-1mm}
\end{table}

Table~\ref{tab-multi-hop_type} presents different types of multi-hop reasonings in our dataset.
Comparison questions require quantitative or logical comparisons between two entities to obtain the answer.
The system is required to understand the properties in the question (e.g., \textit{date of birth}).
Compositional questions require the system to answer several primitive questions and combine them.
For instance, to answer the question \textit{Why did the founder of Versus die?}, the system must answer two sub-questions sequentially: (1) \textit{Who is the founder of Versus?} and (2) \textit{Why did he/she die?}.
Inference questions require that the system understands several logical rules. For instance,  to find the \textit{grandchild}, first, it should find the \textit{child}.
Then, based on the \textit{child}, continue to find the \textit{child}.
Bridge-comparison questions require both finding the bridge entity and doing a comparison to obtain the final answer.

\paragraph{Answer Types}
We preserved all information when generating the data; hence, we used the answer information (both string and Wikidata id) to classify the types of answers.
Based on the value of the property \textit{instance of} in Wikidata, we obtained 708 unique types of answers.
%
The top-5 types of answers in our dataset are: \textit{yes/no} (31.2\%), 
date (16.9\%; e.g., July 10, 2010), 
film (13.5\%; e.g., \textit{La La Land}), 
human (11.7\%; e.g., \textit{George Washington}), 
and big city (4.7\%; e.g., \textit{Chicago}).
%
%
For the remaining types of answers (22.0\%), they are various types of entities in Wikidata.

\section{Experiments}
\label{experiment}

\subsection{Evaluate the Dataset Quality}
\label{equaluate_multihop}
\label{equaluate_quality}
We conducted two different evaluations on our dataset: evaluate the difficulty and the multi-hop reasoning. 
To evaluate the difficulty, we used the multi-hop model as described in~\newcite{data_Hotpotqa} to obtain the results on HotpotQA (distractor setting) and our dataset.
Table~\ref{tab:multi-hop-quality} presents the results.
\myedit{For the SFs prediction task, the scores on our dataset are higher than those on HotpotQA. 
However, for the answer prediction task, the scores on our dataset are lower than those on HotpotQA.
Overall, on the joint metrics, the scores on our dataset are lower than those on HotpotQA.
This indicates that given the human performance on both datasets is comparable (see Section~\ref{sec:human_performance}), the number of difficult questions in our dataset is greater than that in HotpotQA.}

\begin{table}[htp]
  \begin{center}
    \begin{tabular}{l r r r r r r} 
    \Xhline{2\arrayrulewidth}
      \multirow{2}{3cm}{\textbf{Dataset}} & \multicolumn{2}{c}{\textbf{Answer}}  & \multicolumn{2}{c}{\textbf{Sp Fact}}  & 
      \multicolumn{2}{c}{\textbf{Joint}} \\ \cline{2-7}
       & EM & F1 & EM & F1 & EM & F1 \\\hline
      HotpotQA & 44.48 & 58.54 & \textbf{20.68} & \textbf{65.66 }& 10.97 & 40.52 \\
      Our Dataset & \textbf{34.14} & \textbf{40.95} & 26.47 & 66.94 & \textbf{\hphantom{0}9.22} & \textbf{26.76} \\ 
      \Xhline{2\arrayrulewidth}
    \end{tabular}
    \caption{Results (\%) of the multi-hop model on HotpotQA~\protect\cite{data_Hotpotqa} and our dataset. ``\textbf{Sp Fact}'' is the abbreviation for the sentence-level supporting facts prediction task.}
    \label{tab:multi-hop-quality}
  \end{center}
\end{table}

Similar to~\newcite{data_analysis_sewon}, we used a single-hop BERT model~\cite{cite_model_bert} to test the multi-hop reasoning in our dataset.
The F1 score on HotpotQA is 64.6 (67.0 F1 in~\newcite{data_analysis_sewon}); meanwhile, the F1 score on our dataset is 55.9.
The result of our dataset is lower than the result of HotpotQA by 8.7 F1. 
It indicates that a large number of examples in our dataset require multi-hop reasoning to be solved. 
Moreover, it is verified that our data generation and our templates guarantee multi-hop reasoning.
In summary, these results show that our dataset is challenging for multi-hop models and requires multi-hop reasoning to be solved.

\begin{figure}[htp]
    \begin{center}
        \includegraphics[scale=0.39]{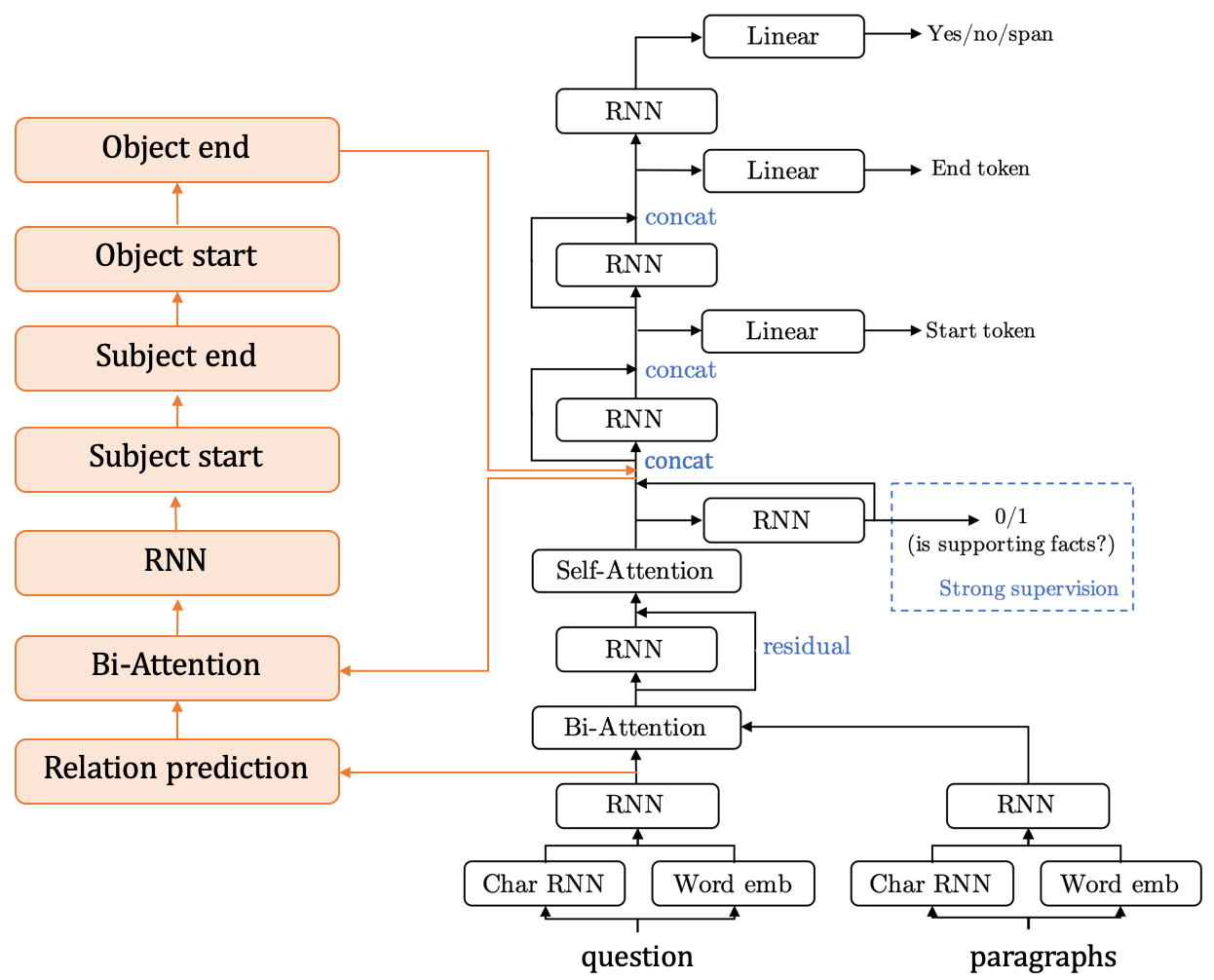}
    \end{center}
    \caption{Our baseline model. The right part is the baseline model of HotpotQA~\protect\cite{data_Hotpotqa}.}
    \label{fig:baseline}
\end{figure}

\subsection{Baseline Results}
We modified the baseline model in~\newcite{data_Hotpotqa} and added a new component (the orange block in Figure~\ref{fig:baseline}) to perform the evidence generation task.
We re-used several techniques of the previous baseline, such as bi-attention, to predict the evidence.
Our evidence information is a set of triples, with each triple including \textit{subject entity}, \textit{relation}, and \textit{object entity}.
First, we used the question to predict the relations and then used the predicted relations and the context (after predicting sentence-level SFs) to obtain the subject and object entities.

Table~\ref{tab:baseline} presents the results of our baseline model.
We used the evaluation metrics as described in Section~\ref{task_evaluation}.
As shown in the table, the scores of the sentence-level SFs prediction task are quite high.
This is a binary classification task that classifies whether each sentence is a SF.
As discussed, this task is incapable of evaluating the reasoning and inference skills of the model.
The scores of the evidence generation task are quite low which indicates this task is difficult.
Our error analysis shows that the model can predict one correct triple in the set of the triples.
However, accurately obtaining the set of triples is extremely challenging.
This is the reason why the EM score is very low.
We believe that adding the evidence generation task is appropriate to test the reasoning and inference skills.

\begin{table}[htp]
  \begin{center}
    \begin{tabular}{l r r r r r r r r} 
    \Xhline{2\arrayrulewidth}
      \multirow{2}{3cm}{\textbf{Split/Task}} & \multicolumn{2}{c}{\textbf{Answer}}  & \multicolumn{2}{c}{\textbf{Sp Fact}}  &
      \multicolumn{2}{c}{\textbf{Evidence}} & 
      \multicolumn{2}{c}{\textbf{Joint}} \\ \cline{2-9}
      
       & EM & F1 & EM & F1 & EM & F1 & EM & F1 \\\hline
      
      Dev & 35.30 & 42.45 & 23.85 & 64.31 & 1.08 & 14.77 & 0.37 & 5.03 \\ 
      Test & 36.53 & 43.93 & 24.99 & 65.26 & 1.07 & 14.94 & 0.35 &  5.41 \\
      \Xhline{2\arrayrulewidth}
    \end{tabular}
    \caption{Results (\%) of the baseline model.}
    \label{tab:baseline}
  \end{center}
\end{table}

To investigate the difficulty of each type of question, we categorized the performance for each type of question (on the test split).
Table~\ref{tab:baseline_ques_type} shows the results.
\myedit{For the answer prediction task, the model obtained high scores on inference and compositional questions.
Meanwhile, for the sentence-level SFs prediction task, the model obtained high scores on comparison and bridge-comparison questions.}
Overall, the joint metric score of the inference question is the lowest. 
This indicates that this type of question is more challenging for the model.
The evidence generation task has the lowest score for all types of questions when compared with the other two tasks.
This suggests that the evidence generation task is challenging for all types of questions.

\begin{table}[htp]
  \begin{center}
    \begin{tabular}{l r r r r r r r r} 
    \Xhline{2\arrayrulewidth}
      \multirow{2}{3cm}{\textbf{Type of Question}} & \multicolumn{2}{c}{\textbf{Answer}}  & \multicolumn{2}{c}{\textbf{Sp Fact}}  &
      \multicolumn{2}{c}{\textbf{Evidence}} & 
      \multicolumn{2}{c}{\textbf{Joint}} \\ \cline{2-9}
      
       & EM & F1 & EM & F1 & EM & F1 & EM & F1 \\\hline
      Comparison & 26.49 & 27.86 & 26.76 & 65.02 & \hphantom{0}0.00 & 12.40 & \hphantom{0}0.00 & 2.45 \\ 
      Inference  & 41.10 & 62.60 & 10.77 & 49.45 & \hphantom{0}0.00 & \hphantom{0}2.85 & \hphantom{0}0.00 & \hphantom{0}1.40 \\  
      Compositional  & 50.40 & 59.94 & 18.28 & 57.44 & \hphantom{0}2.57  & 17.65 & 0.84 & \hphantom{0}9.19 \\
      Bridge-Comparison  & 18.47 & 20.45 &  43.74 & 89.16 & \hphantom{0}0.00 & 19.17 & \hphantom{0}0.00 & \hphantom{0}3.60 \\
      \Xhline{2\arrayrulewidth}
    \end{tabular}
    \caption{Results (\%) of the baseline model on different types of questions.}
    \label{tab:baseline_ques_type}
  \end{center}
\end{table}

\subsection{Human Performance}
\label{sec:human_performance}

We obtained a human performance on 100 samples that are randomly chosen from the test split.
Each sample was annotated by three workers (graduate students).
We provided the question, context, and a set of predefined relations (for the evidence generation task) and asked a worker to provide an answer, a set of sentence-level SFs, and a set of evidence.
Similar to the previous work~\cite{data_Hotpotqa}, we computed the upper bound for human performance by acquiring the maximum EM and F1 for each sample.
All the results are presented in Table~\ref{tab:human}.

\begin{table}[htp]
  \begin{center}
    \begin{tabular}{l r r r r r r r r} 
    \Xhline{2\arrayrulewidth}
      \multirow{2}{3cm}{\textbf{Setting}} & \multicolumn{2}{c}{\textbf{Answer}}  & \multicolumn{2}{c}{\textbf{Sp Fact}}  &
      \multicolumn{2}{c}{\textbf{Evidence}} & 
      \multicolumn{2}{c}{\textbf{Joint}} \\ \cline{2-9}
       & EM & F1 & EM & F1 & EM & F1 & EM & F1 \\\hline
      Model  & 50.00 & 58.48 & 29.00 & 69.90 & \hphantom{0}0.00 & 16.74 & \hphantom{0}0.00 & 9.79 \\ \hline
      Human (average)  & 80.67 & 82.34 & 85.33 & 92.63 & 57.67  & 75.63 & 53.00 & 66.69 \\
      Human Upper Bound (UB) & 91.00 & 91.79 & 88.00 & 93.75 & 64.00  & 78.81  & 62.00 & 75.25 \\
      \Xhline{2\arrayrulewidth}
    \end{tabular}
    \caption{Comparing baseline model performance with human performance (\%) on 100 random samples.}
    \label{tab:human}
  \end{center}
\end{table}


\myedit{
The workers achieved higher performance than that of the model.
The human performance for the answer prediction task is 91.0 EM and 91.8 F1. 
There still seems to be room for improvement, which might be because the mismatch information between Wikipedia and Wikidata makes questions unanswerable (see Section~\ref{error_analysis} for an analysis).
%
The human performance of the answer prediction task on our dataset (91.8 F1 UB) shows a relatively small gap against that on HotpotQA (98.8 F1 UB; borrowed from their paper).
Although the baseline model is able to predict the answer and sentence-level SFs, it is not very effective at finding the evidence.
We also observe that there is a large gap between the performance of human and the model in the evidence generation task (78.8 and 16.7 F1). 
Therefore, this could be a new challenging task for explaining multi-hop reasoning.
We conjecture that the main reason why the score of the evidence generation task was low is the ambiguity in the names of Wikidata.
For example, in Wikidata, one person can have multiple names.
We use only one name in the ground truth, while the workers can use other names.
Future research might explore these issues to ensure the quality of the dataset.
Overall, our baseline results are far behind human performance.
This shows that our dataset is challenging and there is ample room for improvement in the future.}

\subsection{Analysis of Mismatched Examples between Wikipedia and Wikidata}
\label{error_analysis}
\myedit{
As mentioned in Section~\ref{sec:human_performance}, there are unanswerable questions in our dataset due to the mismatch information between Wikipedia articles and Wikidata knowledge.
In the dataset generation process, for a triple $(s, r, o)$, we first checked whether the object entity $o$ appears or not in the Wikipedia article that describes the entity $s$.
Our assumption is that the first sentence in the article in which the object entity $o$ appears is the most important, which we decided to use for the QA pair generation.
For instance, we obtained a triple: \textit{(Lord William Beauclerk, mother, Lady Diana de Vere)} from Wikidata, and we obtained a paragraph $p$ from the Wikipedia article that describes \textit{``Lord William Beauclerk''}.
We used the object entity \textit{``Lady Diana de Vere''} to obtain the first sentence in $p$ \textit{``Beauclerk was the second son of Charles Beauclerk, 1st Duke of St Albans, and his wife Lady Diana de Vere, \ldots.''}
From this sentence, we can infer that the mother of \textit{``Lord William Beauclerk''} is \textit{``Lady Diana de Vere''}.
However, because we only checked whether the object entity $o$ appears in the sentence or not, there could be a semantic mismatch between the sentence and the triple.
For instance, we obtained a triple: \textit{(Rakel Dink, spouse, Hrant Dink)} from Wikidata, while we obtained the first sentence from Wikipedia article: \textit{``Rakel Dink (born 1959) is a Turkish Armenian human rights activist and head of the Hrant Dink Foundation.''}
Obviously, from this sentence, we cannot infer that \textit{``Hrant Dink''} is the spouse of \textit{``Rakel Dink''}.
Therefore, we defined heuristics to exclude these mismatched cases as much as possible.
In particular, we found that some examples have subject entities that are similar/equal to their object entities and are likely to become mismatched cases.
For such cases, we manually checked the samples and decided to use or remove them for our final dataset.
Nonetheless, there are still cases that our heuristics cannot capture.
To estimate how many mismatched cases our heuristics cannot capture in the dataset, we randomly selected 100 samples in the training set and manually checked them.
We obtained eight out of 100 samples that have a mismatch between Wikipedia article and Wikidata triple. 
For the next version of the dataset, we plan to improve our heuristics by building a list of keywords for each relation to check the correspondence between Wikipedia sentence and Wikidata triple.
For instance, we observed that for the relation ``\textit{mother}'', the sentences often contain phrases: ``\textit{son of}'', ``\textit{daughter of}'', ``\textit{his mother}'', and ``\textit{her mother}''.}

\section{Related Work}
\label{related_work}

\paragraph{Multi-hop questions in MRC domain}
Currently, four multi-hop MRC datasets proposed for textual data: ComplexWebQuestions~\cite{data_ComplexWebQuestions},  QAngaroo~\cite{data_Qangaroo}, HotpotQA~\cite{data_Hotpotqa}, and $\mathrm{R^4C}$~\cite{data_r4c}.
%
Recently, \newcite{data_HybridQA} introduced the HybridQA dataset---a multi-hop question answering over both tabular and textual data.
The dataset was created by crowdsourcing based on Wikipedia tables and Wikipedia articles.

\paragraph{Multi-hop questions in KB domain}
Question answering over the knowledge graph has been investigated for decades.
However, most current datasets~\cite{data_webquestion,data_simplequestion,data_webquestion_sp,data_simplequestion_wiki} consist of simple questions (single-hop).
~\newcite{qa_kb_zhang} introduced the METAQA dataset that contains both single-hop and multi-hop questions.
\newcite{kg_data_1} introduced the ComplexQuestions dataset comprising 150 compositional questions.
All of these datasets are solved by using the KB only.
Our dataset is constructed based on the intersection between Wikipedia and Wikidata. 
Therefore, it can be solved by using structured or unstructured data.

\paragraph{Compositional Knowledge Base Inference}
Extracting Horn rules from the KB has been
studied extensively in the Inductive Logic Programming literature~\cite{rule_ilp,rule_ilp2}.
From the KB, there are several approaches that mine association rules~\cite{associate_rule} and several mine
logical rules~\cite{rule_sherlock,logical_rule_amie}.
We observed that these rules can be used to test the reasoning skill of the model.
Therefore, in this study, we utilized the logical rules in the form: $r_1(a, b) \wedge r_2(b, c) \Rightarrow r(a, c)$.
ComplexWebQuestions and QAngaroo datasets are also utilized KB when constructing the dataset, 
but they do not utilize the logical rules as we did.

\paragraph{RC datasets with explanations}
Table~\ref{tab:data_explanation} presents several existing datasets that provide explanations.
HotpotQA and $\mathrm{R^4C}$ are the most similar works to ours.
HotpotQA provides a justification explanation (collections of evidence to support the decision) in the form of a set of sentence-level SFs.
$\mathrm{R^4C}$ provides both justification and introspective explanations (how a decision is made). 
Our study also provides both justification and introspective explanations.
The difference is that the explanation in our dataset is a set of triples, where each triple is a structured data obtained from Wikidata.
Meanwhile, the explanation in $\mathrm{R^4C}$ is a set of semi-structured data.
$\mathrm{R^4C}$ is created based on HotpotQA
and has 4,588 questions.
The small size of the dataset implies that it cannot be used for training end-to-end neural network models involving the multi-hop reasoning with comprehensive explanation.

\begin{table}[htp]
  \begin{center}
    \begin{tabular}{l c c r}  
    \Xhline{2\arrayrulewidth}
      \multirow{2}{3cm}{\textbf{Task/Dataset}} & \multicolumn{2}{c}{\textbf{Explanations}} & \multirow{2}{1cm}{\textbf{Size}} \\
       & Justification &  Introspective &  \\ \hline
      Our work & \checkmark & \checkmark & 192,606 \\ \hline
      
      $\mathrm{R^4C}$~\cite{data_r4c} & \checkmark & \checkmark & 4,588 \\
      
      CoS-E~\cite{data_rajani} &  & \checkmark & 19,522 \\
      
      HotpotQA~\cite{data_Hotpotqa} & \checkmark &  & 112,779 \\
      
      Science Exam QA~\cite{data_jansen} &  & \checkmark & 363 \\ 
      
    \Xhline{2\arrayrulewidth}
    \end{tabular}
    \caption{Comparison with other datasets with explanations.}
    \label{tab:data_explanation}
  \end{center}
\end{table}

\section{Conclusion}
\label{conclustion}
In this study, we presented 2WikiMultiHopQA---a large and high quality multi-hop dataset that provides comprehensive explanations for predictions.
We utilized logical rules in the KB to create more natural questions that still require multi-hop reasoning.
Through experiments, we demonstrated that our dataset ensures multi-hop reasoning while being challenging for the multi-hop models.
We also demonstrated that bootstrapping the multi-hop MRC dataset is beneficial by utilizing large-scale available data on Wikipedia and Wikidata.


\section*{Acknowledgments}
\myedit{
We would like to thank An Tuan Dao, Johannes Mario Meissner Blanco, Kazutoshi Shinoda, Napat Thumwanit, Taichi Iki, Thanakrit Julavanich, and Vitou Phy for their valuable support in the procedure of constructing the dataset.
We thank the anonymous reviewers for suggestions on how to improve the dataset and the paper. 
This work was supported by JSPS KAKENHI Grant Number 18H03297.}

\bibliographystyle{coling}
\bibliography{coling2020}

\newpage 
\appendix

\section{Data Collection Details}
\label{sec:data_collection_details}

\subsection{Data Preprocessing}
\label{appendix_data_process}
We used both dump\footnote{\url{https://dumps.wikimedia.org/}} and online version of Wikipedia and Wikidata.
We downloaded the dump of English Wikipedia on January 1, 2020, and the dump of English Wikidata on December 31, 2019.
From Wikidata and Wikipedia, we obtained 5,950,475 entities.
Based on the value of the property \textit{instance of} in Wikidata, we categorized all entities into 23,763 groups.
In this dataset, we focused on the most popular entities (top-50 for comparison questions).
When checking the requirements to ensure the multi-hop reasoning of the dataset, several entities in the multi-path are not present in the dump version; in such situations, we used the online version of Wikipedia and Wikidata.

We observed that the quality of the dataset depends on the quality of the intersection information between Wikipedia and Wikidata.
Specifically, for the property related to date information, such as \textit{publication date} and \textit{date of birth}, information between Wikipedia and Wikidata is quite consistent.
Meanwhile, for the property \textit{occupation}, information between Wikipedia and Wikidata is inconsistent. 
For instance, the Wikipedia of the entity \textit{Ebenezer Adam} is as follows: ``\textit{Ebenezer Adam was a Ghanaian educationist and politician.}''; meanwhile, the value from Wikidata of the property \textit{occupation} is \textit{politician}.
In such situations, we manually check all samples related to the property to ensure dataset quality.
For the property related to the country name, we handled many different similar names by using the aliases of the entity and the set of demonyms.
Moreover, to guarantee the quality of the dataset, we only focused on the set of properties with high consistency between Wikipedia and Wikidata.

We used both Stanford CoreNLP~\cite{sentence_segmentation} and Spacy to perform sentence segmentation for the context.

\subsection{Comparison Questions}
\label{appendix_comparison}
Table~\ref{tab:template_info_compare} presents all information of our comparison question.
We can use more entities and properties from Wikidata to create a dataset.
In this version of the dataset, we focused on the top-50 popular entities in Wikipedia and Wikidata. 
To ensure dataset quality, we used the set of properties as described in the table.
For each combination between the entity and the property, we have various templates for asking questions to ensure diversity in the questions.

\begin{table}[htp]
  \begin{center}
    \begin{tabular}{{p{8cm} p{4cm} p{2cm}}}  
    \Xhline{2\arrayrulewidth}
      Entity Type & Property & \#Templates \\\hline
      Human & date of birth & 7\\ 
      & date of death & 3\\ 
      & date of birth and date of death (year old) & 2 \\
      & occupation & 18\\
      & country of citizenship & 11\\
      & place of birth & 1\\ \hline
       
      Film & publication date & 5\\ 
      & director & 2\\ 
      & producer & 2\\ 
      & country of origin & 7 \\ \hline
      
      Album & publication date & 5\\ 
      & producer & 2\\ \hline
      
    Musical group & inception & 4\\ 
      & country of origin & 7\\ \hline
      
    Song & publication date & 5\\ \hline
      
    \multirow{2}{7cm}{Museum, Airport, Magazine, Railway station, Business, Building, Church building, High school, School, University}   
    
     & inception & 1-3 \\
     & country & 4 \\ \hline
     
    Mountain, River, Island, Lake, Village & country & 4\\ 
    \Xhline{2\arrayrulewidth}
    \end{tabular}
    \caption{Templates of Comparison questions.}
    \label{tab:template_info_compare}
  \end{center}
\end{table}

\subsection{Inference Questions}
\label{appendix_inference}
We argued that logical rules are difficult to apply to multi-hop questions.
We obtained a set of 50 inference relations, but we cannot use all of it into the dataset.
For instance, the logical rule is $place of birth(a, b) \wedge country(b, c) \Rightarrow nationality(a, c)$; this rule easily fails after checking the requirements.
To guarantee the multi-hop reasoning of the question, the document describing a person $a$ having a place of birth $b$ should not contain the information about the country $c$.
However, most paragraphs describing humans often contain information on their nationality.

The other issue is ensuring that each sample has only one correct answer on the two gold paragraphs.
With the logical rule being $child(a, b) \wedge child(b, c) \Rightarrow grandchild(a, c)$, if $a$ has more than one child, for instance $a$ has three children $b_1$, $b_2$ and $b_3$,
then each $b$ has their own children.
Therefore, for the question ``\textit{Who is the grandchild of $a$?}'', there are several possible answers to this question.
To address this issue in our dataset, we only utilized the relation that has only one value in the triple on Wikidata.
That is the reason why the number of inference questions in our dataset is quite small.
Table~\ref{tab:inference_info} describes all inference relations used in our dataset.

In most cases, this rule will be correct.
However, several rules can be false in some cases.
In such situations, based on the Wikidata information, we double-checked the new triple before deciding whether to use it.
For instance, the rule is
$doctoral\_advisor(a, b) \wedge employer(b, c) \Rightarrow educated\_at(a, c)$, $a$ has an advisor is $b$, $b$ works at $c$, and we can infer that $a$ studies at $c$.
There can be exceptions that $b$ works at many places, and $c$ is one of them, but $a$ does not study at $c$.
We used Wikidata to check whether $a$ studies at $c$ before deciding to use it.

To obtain the question, we used the set of templates in Table~\ref{tab:inference_info_template_ques}.

\begin{table}[htp]
  \begin{center}
    \begin{tabular}{{p{4cm} p{4cm} p{4cm}}}  
     \Xhline{2\arrayrulewidth}
      Relation 1 & Relation 2 & Inference Relation  \\\hline
      spouse & spouse & co-husband/co-wife  \\
      spouse & father & father-in-law  \\
      spouse & mother & mother-in-law  \\
      spouse & sibling & sibling-in-law \\
      spouse & child & child/stepchild \\ \hline

      father & father & paternal grandfather \\
      father & mother & paternal grandmother  \\
      father & spouse & mother/stepmother  \\
      father & child & sibling \\
      father & sibling & uncle/aunt \\ \hline
      
      mother & mother & maternal grandmother  \\
      mother & father & maternal grandfather  \\
      mother & spouse & father/stepfather  \\
      mother & child & sibling  \\
      mother & sibling & uncle/aunt \\ \hline
      
      child & child & grandchild  \\
      child & sibling & child  \\
      child & mother & wife  \\
      child & father & husband  \\
      child & spouse & child-in-law  \\ \hline
      
      sibling & sibling & sibling \\
      sibling & spouse & sibling-in-law  \\
      sibling & mother & mother \\
      sibling & father & father  \\ \hline
      
      doctoral student & educated at & employer  \\
      doctoral student & field of work & field of work  \\
      doctoral advisor & employer & educated at  \\
      doctoral advisor & field of work & field of work  \\
     \Xhline{2\arrayrulewidth}
    \end{tabular}
    \caption{Inference relation information in our dataset.}
    \label{tab:inference_info}
  \end{center}
\end{table}

\begin{table}[htp]
  \begin{center}
    \begin{tabular}{{p{8.5cm} p{6cm}}} 
    \Xhline{2\arrayrulewidth}
      Relation & Template(s) \\\hline
      \multirow{7}{7cm}{aunt, child-in-law, child, co-husband, co-wife, father-in-law, father, grandchild, grandfather,  grandmother, husband, mother-in-law,
      mother, sibling-in-law, sibling, stepchild,
      stepfather, stepmother, uncle, wife}
      & \\ 
      & \\ 
      & Who is the \#relation of \#name? \\
      & Who is \#name's \#relation? \\ 
      & \\ 
      & \\ 
      & \\ 
      \hline
      educated at & Which \#instance\_of\_answer did \#name study at? \\
      & Which \#instance\_of\_answer did \#name graduate from? \\
      \hline
      employer & Which \#instance\_of\_answer does \#name work at? \\
      & Where does \#name work? \\ \hline
      field of study & What is the field of study of \#name? \\
    \Xhline{2\arrayrulewidth}
    \end{tabular}
    \caption{Templates of Inference question.}
    \label{tab:inference_info_template_ques}
  \end{center}
\end{table}

\subsection{Compositional Questions}
\label{appendix_composite}
For this type of question, we utilized various entities and properties on Wikidata.
We used the following properties (13 properties) as the first relation:
\textit{composer, creator, director, editor, father, founded by, has part, manufacturer, mother, performer, presenter, producer, and spouse}.
Further, we used the following properties (22 properties) as the second relation: \textit{date of birth, date of death, place of birth, country of citizenship, place of death, cause of death, spouse, occupation, educated at, award received, father, place of burial, child, employer, religion, field of work, mother, inception, country, founded by, student of, and place of detention}.
A compositional question was created by combining the first relation and the second relation (ignore duplicate case).

We used the following entities (15 entities) to create this type of question: \textit{human, film, animated feature film, album, university, film production company, business, television program, candy, written work, literary work, musical group, song, magazine, newspaper}.
We obtained a total of 799 templates.

\clearpage 
\subsection{Bridge-comparison Questions}
\label{appendix_combine}
The top-3 popular entities on Wikipedia and Wikidata are \textit{human}, \textit{taxon}, and \textit{film}.
In this type of question, we focused on the combination between \textit{human} and \textit{film}.
Table~\ref{tab:combine_info} presents
the combination between the relations from the two entities \textit{human} and \textit{film} in our dataset.

\begin{table}[htp]
  \begin{center}
    \begin{tabular}{{p{4cm} p{4cm}}}  
    \Xhline{2\arrayrulewidth}
      Relation 1 & Relation 2 \\\hline
      director & date of birth \\
      director & date of death \\
      director & country of citizenship \\ \hline
      producer & date of birth \\
      producer & date of death \\
      producer & country of citizenship \\
      \Xhline{2\arrayrulewidth}
    \end{tabular}
    \caption{Bridge-comparison question's information.}
    \label{tab:combine_info}
  \end{center}
\end{table}
For each row in Table~\ref{tab:combine_info}, we have several ways to ask a question.
For instance, in the first row, with the combination of the two relations \textit{director} and \textit{date of birth}, we have various ways to ask a question, as shown in Table~\ref{tab:combine_template}.
To avoid ambiguous cases, we ensured that each film we used has only one director or one producer.
A total of 62 templates was obtained for this type of question.

\begin{table}[htp]
  \begin{center}
    \begin{tabular}{l}  
    \Xhline{2\arrayrulewidth}
     Templates \\\hline
Which film has the director born first, \#name or \#name? \\
Which film whose director was born first, \#name or \#name? \\
Which film has the director who was born first, \#name or \#name? \\
Which film has the director born earlier, \#name or \#name? \\
Which film has the director who was born earlier, \#name or \#name? \\
Which film whose director is younger, \#name or \#name? \\
Which film has the director born later, \#name or \#name? \\
Which film has the director who was born later, \#name or \#name? \\
Which film has the director who is older than the other, \#name or \#name? \\
Which film has the director who is older, \#name or \#name? \\
        \Xhline{2\arrayrulewidth}
    \end{tabular}
    \caption{Templates of Bridge-comparison questions.}
    
    \label{tab:combine_template}
  \end{center}
\end{table}

\subsection{Generate Data}
\label{generate_data}
\myedit{The algorithms for generating comparison questions and bridge questions are described in Algorithm~\ref{al:generate_compare_data} and Algorithm~\ref{al:generate_brdige_data}, respectively.}

\begin{algorithm}[htp]
\SetAlgoLined
\KwIn{Set of all templates, all entities in the same group, Wikipedia and Wikidata information for each entity}
\KwOut{A question--answer pair with these information: question $Q$, answer $A$, context $C$, sentence-level SFs $SF$, and evidences $E$}
 \While{$\mathrm{not~finished}$}{
  Randomly choose two entities $e_1$ and $e_2$\;
  Obtain all triples (relations and objects) of each entity from Wikidata\;
  Obtain a set of mutual relations ($M$) between two entities\;
  Obtain Wikipedia information of each entity\;
  \For{$\mathrm{each~relation~in~} M$}{
    \If{$\mathrm{pass~requirements}$}{
     Choose a template randomly\;
     Generate a question $Q$\;
     Obtain a context $C$\;
     Obtain an evidence $E$\;
     Compute an answer $A$\;
     Compute sentence-level SFs $SF$\;
    }
  }
  }
 \caption{Comparison Question Generation Procedure}
 \label{al:generate_compare_data}
\end{algorithm}

\begin{algorithm}
\SetAlgoLined
\KwIn{Set of relations $R$, Wikipedia and Wikidata information for each entity}
\KwOut{A question--answer pair with these information: question $Q$, answer $A$, context $C$, sentence-level SFs $SF$, evidences $E$}
 \While{$\mathrm{not~finished}$}{
  Randomly choose an entity $e$\;
  Obtain a set of statements (relations and objects) of the entity from Wikidata\;
  Filter the set of statements based on the first relation information in $R$ to obtain a set of 1-hop  $H_1$\;
  For each element in $H_1$, do the same process (from Line 3) to obtain a set of 2-hop $H_2$,  each element in $H_2$ is a tuple $(e, r_1, e_1, r_2, e_2)$\;
  \For{$\mathrm{each~tuple~in~} H_2$}{
    Obtain Wikipedia articles for two entities: $e$ and $e_1$\;
    \If{$\mathrm{pass~requirements}$ }{
     Choose a template randomly based on $r_1$ and $r_2$\;
     Generate a question $Q$\;
     Obtain a context $C$\;
     Obtain an evidence $E$\;
     Obtain an answer $A$\;
     Compute sentence-level SFs $SF$\;
    }
  }
  }
 \caption{Bridge Question Generation Procedure}
 \label{al:generate_brdige_data}
\end{algorithm}

\subsection{Post-process Generated Data}
\label{post_process_data}
For the bridge questions, we created the data from the two triples $(e, r_1, e_1)$ and $(e_1, r_2, e_2)$.
When we have another triple $(e, r_1, e_{1*})$ that has the same entity and the property with the first triple, it becomes an ambiguous case.
Hence, we discarded all such cases in our dataset based on the information from Wikidata.

For the comparison questions, when a question is asked for comparing two entities about numerical values and the values of the two entities are equal, we remove it.

\end{document}